%% file: main.tex
\documentclass{article}
\usepackage[utf8]{inputenc} 
\usepackage[T1]{fontenc}    
\usepackage{hyperref}       
\usepackage{url}            
\usepackage{booktabs}       
\usepackage{amsfonts}       
\usepackage{nicefrac}       
\usepackage{microtype}      
\usepackage{xcolor}         
\usepackage{graphicx}
\usepackage{subcaption}
\usepackage{amsmath}
\input{package_defs}
\usepackage{geometry}
\title{A Reproducible Extraction of Training Images from Diffusion Models}
\author{%
Ryan Webster\\
Unicaen\\
{\tt\small ryan.webster@unicaen.fr}
}

\begin{document}

\maketitle
\begin{abstract}
    Recently, \cite{carlini2023extracting} demonstrated the widely used model Stable Diffusion can regurgitate real training samples, which is troublesome from a copyright perspective. In this work, we provide an efficient extraction attack on par with the recent attack, with several order of magnitudes less network evaluations. In the process, we expose a new phenomena, which we dub template verbatims, wherein a diffusion model will regurgitate a training sample largely in tact. Template verbatims are harder to detect as they require retrieval and masking to correctly label. Furthermore, they are still generated by newer systems, even those which de-duplicate their training set, and we give insight into why they still appear during generation. We extract training images from several state of the art systems, including Stable Diffusion 2.0, Deep Image Floyd, and finally Midjourney v4. We release code to verify our extraction attack, perform the attack, as well as all extracted prompts at \url{https://github.com/ryanwebster90/onestep-extraction}.
\end{abstract}

\input{intro}

\input{attack_model}

\input{results}

\bibliographystyle{ieee_fullname}
\bibliography{bib}

\end{document}

%% file: package_defs.tex
\usepackage{array}
\usepackage{subcaption}
\usepackage{multirow}
\usepackage{multicol}
\usepackage{tabularx}
\usepackage{bbm}


%% file: intro.tex
\section{Introduction}


Over the past few years, advances in large scale image generation systems have been brought on by publicly available billions scale datasets \cite{schuhmann2021laion,schuhmann2022laion,stablediffusion2022,rombach2022high}. Due to their high quality, generality and ease of use, image generation systems such as Midjourney \cite{midjourney} and Stable Diffusion \cite{stablediffusion2022} have garnished millions of activate users, most of which have little technical knowledge. To train such models, largely automated bots search for suitable text and image pairs from all over the web, as is the case with the training set of Stable Diffusion, LAION-2B \cite{schuhmann2022laion}. 
\par{}
The widespread use of these generation systems is a testament to their generalization capacity; users often want to generate an image which doesn't exist or transform existing images to add new content. However, recent demonstrations show that popular systems such as Stable Diffusion can regurgitate exact copies of training images \cite{openaipremitigate,carlini2023extracting,somepalli2022diffusion,webster2023deduplication}. In \cite{openaipremitigate,carlini2023extracting}, highly duplicated images appear to be a necessary but not sufficient precursor to model memorization. Also, several new works suggest the training datasets do have a high level of duplication \cite{webster2023deduplication,abbas2023semdedup}, with \cite{webster2023deduplication} showing that almost a third of the two billion images in the Stable Diffusion v1 are near duplicates. 
\par{}

Whilst democratizing the creative process is a promising feature of generative models, they also can negatively and unfairly impact existing artists. This is especially true for artists whose work was scraped during dataset creation and who are not correctly attributed. However, enforcing copyright in the U.S. is often a difficult, complicated and expensive battle \cite{netanel2011making}, not to mention the law is largely undefined for generative models \cite{franceschelli2021copyright}. For Stable Diffusion, it's fairly easy to prove your training data was used, as the training set is known \cite{gettylawsuit}, but for closed source systems, one would have to employ a membership inference attack, which is a challenging problem in its own right \cite{hayes2018logan,carlini2023extracting}.
Thus, we focus on reproducibility in this work; releasing both the code and an efficient method running on consumer machines.
\par{}
In this work, we provide the following contributions:
\begin{itemize}
    \item In Sec.~\ref{sec:dcs_score}, we observe that popular models can regurgitate samples with a single sampling step. We use this fact to construct whitebox and blackbox attacks. We also describe \textit{template verbatims}, wherein a training sample is copied largely in tact, with non-semantic variations in fixed image locations.
    \item In Sec.~\ref{sec:results}, we show that our attack is largely on par with the one in \cite{carlini2023extracting}, whilst taking considerably less network evaluations. It can also be used in tandem with \cite{carlini2023extracting} as a post filtering. 
    \item Finally, in Sec.~\ref{sec:ddpm_attack_insight}, we extract samples from a variety of other models, including the closed source system Midjourney. We also provide several insights into the nature of template verbatims and why they still appear even in systems that deduplicate their training set.
\end{itemize}


\section{Related work}

\paragraph{Billions Scale Datasets}

If the first "web scale" image dataset, LAION-400M, was released just a few years ago \cite{schuhmann2021laion}, even larger datasets have been subsequently released \cite{kakaobrain2022coyo-700m,schuhmann2022laion}. The most widely used is the LAION-5B dataset, with roughly 5 billion text image pairs and the corresponding english language subset LAION-2B-en \cite{schuhmann2022laion}. These datasets are automatically collect and text image/text pairs are only selected if they have a high enough CLIP score \cite{radford2021learning}. The CLIP network \cite{radford2021learning,OpenCLIP}, is trained to align image and text features with a contrastive loss and can provide a score that corresponds to a caption's relevance to an image. Thus, during construction of LAION-5B CLIP features are computed and the authors released CLIP features alongside the dataset. 
\paragraph{Deduplication}
In \cite{webster2023deduplication}, the CLIP image features were used to deduplicate LAION-2B, which revealed almost a third of the dataset are near duplicates. In fact, in subsequent work, we noted that around 500M of these are MD5 duplicates; i.e. duplicates at the file level. In \cite{carlini2023extracting,openaipremitigate}, it was noted that highly duplicated samples tend to be memorized by diffusion models and in \cite{openaipremitigate}, they de-duplicated before training. Likewise, in Stable Diffusion 2.0 \cite{stablediffusion2022}, the author's used a perceptual hash to deduplicate before training. As we'll see in Sec.~\ref{sec:transfer_attack}, this is effective at mitigating the verbatims found in \cite{carlini2023extracting}, but not all. Finally, in SemDeDup \cite{abbas2023semdedup}, it was noted that around 50\% of the samples of LAION-400M, which have a very close semantic pair, can simply be removed while keeping or improving the zero-shot performance of CLIP.
\paragraph{Membership Inference}
As we touched on in the introduction, in order to enforce copyright, owners would have to first prove their data exists within the models training set. For models like Midjourney or DALLE-2, the model parameters and training sets are unknown. In machine learning, the process of determining which training samples were used only from model outputs, is known as \textit{membership inference} and in particular the challenging "black box" setting. Membership inference has been widely studied against GAN generated images, with varying success, for a wide variety of different settings, for instance in \cite{hayes2018logan}. Several very recent approaches have designed black box attacks specifically against diffusion models \cite{matsumoto2023membership,duan2023diffusion,hu2023membership}, but conduct studies versus relatively small datasets (<1M samples). Namely, in the recent work in \cite{hu2023membership} a loss based attack is presented which bears resemblance to our whitebox attack (See Eq.~\ref{eq:ddpm_attack_dcs}), however they do so in the unconditional case, and observe better performance later in the generation process.
\input{figures/fig_attack_description}
\paragraph{Extraction Attacks}
A special case of membership inference, is the ability to actually reconstruct training samples from model outputs \cite{carlini2021extracting,carlini2023extracting}. Extraction is a much harder problem and typically is only possible for very few samples. 
In \cite{carlini2023extracting}, only roughly 100 images were reconstructed from stable diffusion, out of 350K attempted prompts. Thus, this setting typically is only concerned with precision, and number of samples extracted, rather than the typical precision recall in membership inference. Extracted samples however clearly have higher implications to privacy or copyright.

%% file: figures/fig_attack_description.tex
\begin{figure}[!ht]
    \centering
    \small
    \renewcommand{\arraystretch}{1.2} 
    \begin{tabular}{>{\centering\arraybackslash}m{0.13\linewidth}m{0.13\linewidth}m{0.13\linewidth}m{0.13\linewidth}m{0.13\linewidth}m{0.13\linewidth}}
        \multicolumn{1}{c}{\textbf{Type}} & \multicolumn{1}{c}{\textbf{One Step}} & \multicolumn{1}{c}{\textbf{New Noise}} & \multicolumn{1}{c}{\textbf{Full Step}} & \multicolumn{1}{c}{\textbf{Match}} & \multicolumn{1}{c}{\textbf{Retrieve}} \\ 
        \centering Exact Verbatim  &
            \includegraphics[width=\linewidth]{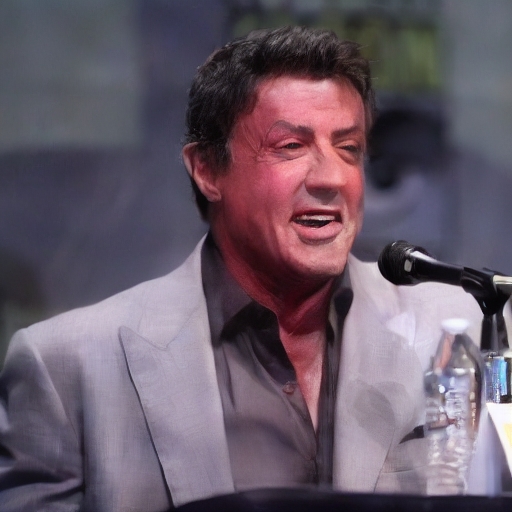} &
            \includegraphics[width=\linewidth]{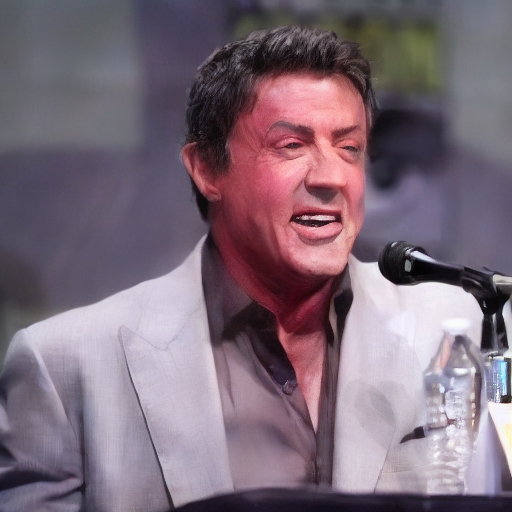} &

            \includegraphics[width=\linewidth]{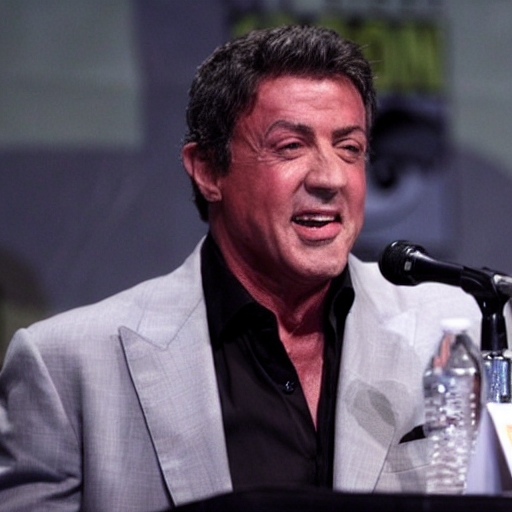} &
            \includegraphics[width=\linewidth]{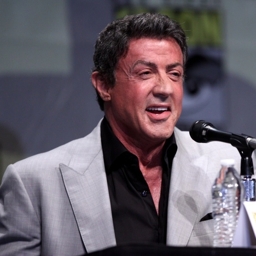}
            &
            \includegraphics[width=\linewidth]{figures/fig_attack_description/rambo_real.jpeg}
            \\
        \centering Template Verbatim  &
            \includegraphics[width=\linewidth]{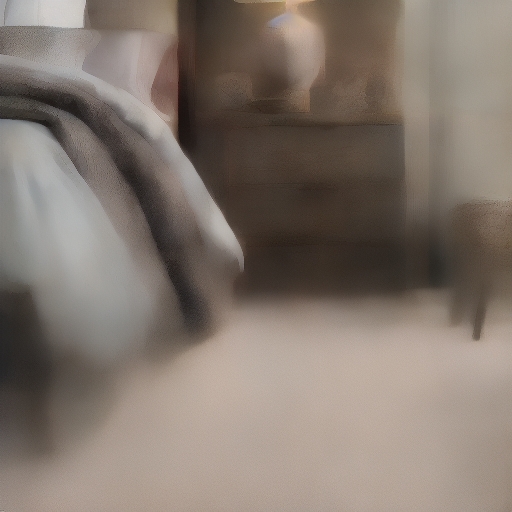} &
            \includegraphics[width=\linewidth]{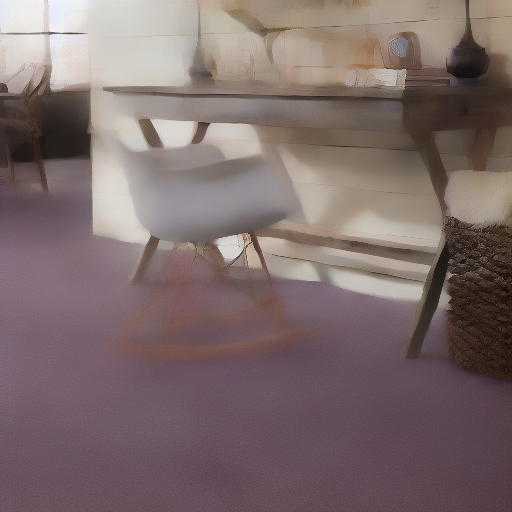} &
            \includegraphics[width=\linewidth]{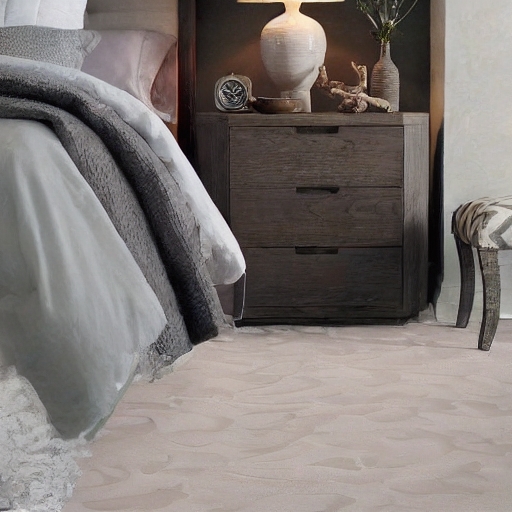} &
            \includegraphics[width=\linewidth]{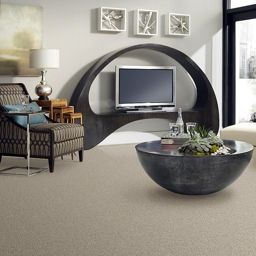}&
            \includegraphics[width=\linewidth]{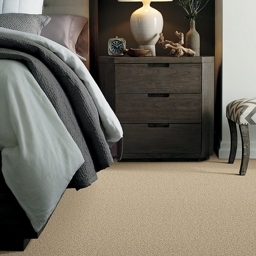}
            \\
        \centering Non-Verbatim  &
            \includegraphics[width=\linewidth]{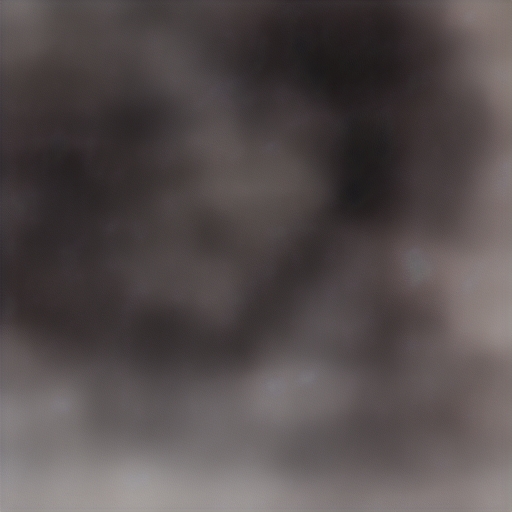} &
            \includegraphics[width=\linewidth]{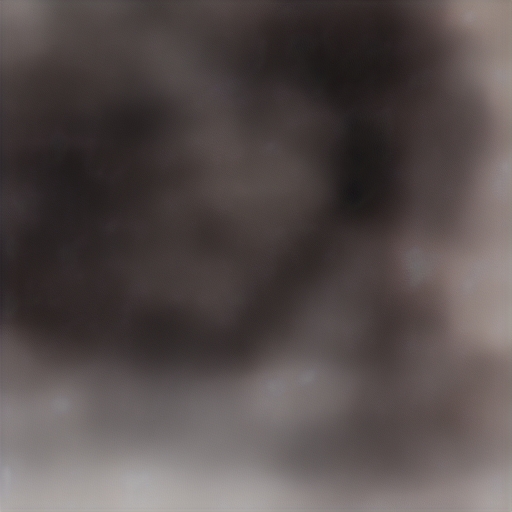} &
            \includegraphics[width=\linewidth]{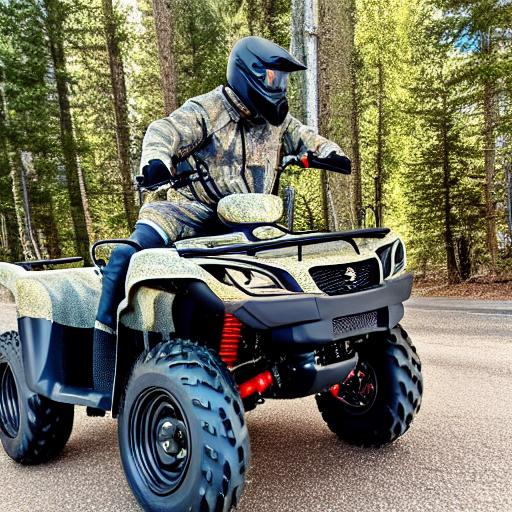} &
            \includegraphics[width=\linewidth]{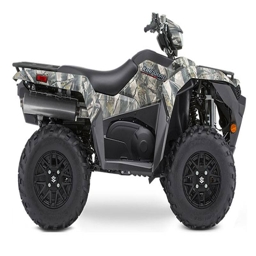}
            &
            \includegraphics[width=\linewidth]{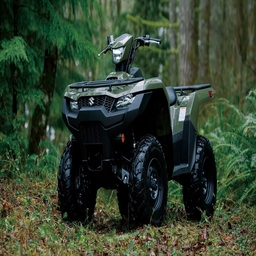}
            \\
    \end{tabular}
    \caption{Training images can be extracted from Stable-Diffusion in one step. In the first row, a verbatim copy is synthesized from the caption corresponding to the image on the second to last column. In the second row, we present verbatim copies that are harder to detect: \textit{template verbatims}. They typically represent many-to-many mappings (many captions synthesize many verbatim templates) and thus the ground truth is constructed with retrieval (right most column). Furthermore, they exhibit variation in fixed locations of the image; here the color of the carpet can change so they must be masked to be correctly detected. Non-verbatims have no match, even when retrieving over the entire dataset. See Sec.~\ref{sec:ddpm_attack_rv} for how we construct the ground truth here in this case.}
    \label{fig:ddpm_attack_teaser}
\end{figure}

%% file: attack_model.tex
\section{Attack Model}\label{sec:dcs_score}
In our initial experiments, we reproduced a small scale version of \cite{carlini2023extracting}. We used the open source de-duplication in \cite{webster2023deduplication}, which finds near duplicates using compressed CLIP image features and synthesized a small set of the most duplicated images. We found verbatim copies have the property that they can be synthesized in a single iteration. Fig.~\ref{fig:ddpm_attack_teaser} shows clearly different behaviors for prompts that are copied versus those that are not. We use this as an intuition to construct a whitebox attack presented next.

\subsection{Whitebox Attack}
In this setting, we assume the attacker has both the captions and the model parameters. 
Recall that conditional diffusion models train a denoising autoencoder $D(z_{t},c)$, which will recover the real sample $z_{0}$ given its noised version $z_{t}$. At synthesis time, a sample is drawn from a gaussian distribution $\epsilon \sim N(0,1)$ and given a variance $\sigma_{T}$ for timesteps $T$ depending on the sampler (see \cite{rombach2022high,karras2020analyzing} for more details). We propose an easy to compute metric, which can capture the one step synthesis property in Fig.~\ref{fig:ddpm_attack_teaser} by measuring how much $D(\sigma_{T} \epsilon,c)$ modifies the noise as follows:

\begin{equation}
    \mathrm{DCS}(c) := \left\Vert \sigma_{1}\epsilon - D(\sigma_{1}\epsilon,c) \right\Vert^{2}_{2}
\end{equation}\label{eq:ddpm_attack_dcs}

Here, $\sigma_{1}$ is initialized given the Heun sampler implemented in \cite{kdiffusiongh} with one time step. We call this error the Denoising Confidence Score (DCS). we can turn this into a binary classifier then by threshing it's values with $\tau_{\mathrm{DCS}}$ as follows
\begin{equation}
    F_{\mathrm{DCS}}(c;\tau) = \mathbbm{1}_{\mathrm{DCS}(c) \geq \tau_{\mathrm{DCS}}}
\end{equation}\label{eq:ddpm_attack_dcs_classifier}

Whilst this is "whitebox," we in fact don't need access to the parameters of the diffusion model, just the noise generation process. For latent diffusion models, we just require the first stage downsampling encoder.

\subsection{Black Box Setting}
We consider the setting similar to \cite{carlini2023extracting}, wherein an attacker has captions but can only invoke $\texttt{Gen}(c,r)$ for caption $c$, random seed $r$ and we also assume the ability to control the timesteps $T$ (as is the case for Midjourney). Henceforth, we use $\texttt{Gen}(c,r,T=1)$ when we control the timesteps and simply $\texttt{Gen}(c,r)$ for full synthesis.
We construct our attack inspired by the observation in Fig~\ref{fig:ddpm_attack_teaser}, that is we still use one iteration for synthesis. We noticed non verbatims normally start as highly blurry images (Fig~\ref{fig:ddpm_attack_teaser}, last row). Blurry images will not have consistent edges (or no edges at all) in contrast to the natural appearing verbatims with clear edges. We thus employ a simple image processing technique by first computing the images edges, and then look for how consistently these same edges appear for other seeds. Letting the operator $\texttt{Edge}()$ outputting a binary image from threshing the magnitude of spatial gradients (e.g. with a sobel filter), we define the Edge Consistency Score (ECS) as follows
\begin{equation}\label{eq:ddpm_attack_ecs}
L_{\mathrm{ECS}}(c) = \left\Vert \left( \frac{1}{J} \sum_{j\leq J} \texttt{Edge}(\texttt{Gen}(c,r_{j},T=1)) \right ) \geq \gamma \right\Vert ^{2}_{2}
\end{equation}
Where $J$ represents generating over several random seeds and $\gamma$ a threshhold. Note that the term on the LHS of Eq.~\ref{eq:ddpm_attack_ecs} is the average image of binary edges, thus $\gamma$ should be chosen proportional to $J$. We construct a binary classifier in the same way as in Eq.~\ref{eq:ddpm_attack_dcs_classifier}. Whilst crude, we find that this score works decent in practice.

\input{figures/example_templates}

\section{Constructing a Ground Truth}\label{sec:ddpm_attack_constructing_gt}
As we'd like to evaluate the precision of the above attacks, we need to ascertain images that are truly copied by the model. We describe two ways to label images as copied below:

\paragraph{Matching Verbatims (MV)}
In \cite{carlini2023extracting}, an image is considered to be a ground truth verbatim copy if its MSE between the generated image from caption $c$ and it's corresponding dataset image $x$ is small enough, i.e.
\begin{equation}
    L_{MV}(x,c;J) = \min_{j \leq J}\left\Vert x - \texttt{Gen}(c,r_{j}) \right \Vert ^{2}_{2} 
\end{equation}\label{eq:ddpm_attack_verb_mv}
We take the minimum over $J$ random seeds $r_{j}$ as sometimes verbatims appear after a few seeds. Then, the ground truth labels are simply a thresh-hold on this distance
\begin{equation}
\texttt{IsVerb}(x,c;J,\delta_{V}) = \mathbbm{1}_{L_{MV}(x,c;J) \leq \delta_{V}}
\end{equation}
We chose $\delta_{V} = .12$. This is slightly more relaxed than what was chosen in \cite{carlini2023extracting}, and simply prune false positives by hand (such as images of textures or without objects that can be false positives), which given the rarity of positively labeled images is feasible.
\paragraph{Retrieval Verbatims}\label{sec:ddpm_attack_rv}
We noticed that some images would not correspond to their matching image, despite having the curious property of one-step synthesis (see Fig.~\ref{fig:ddpm_attack_teaser}). Thus, we retrieved the images with an index \cite{webster2023deduplication,clipretrieval}. Furthermore, we found that many images which had very few duplicates, had extremely many near duplicates that differed in only one region of the image. These images would commonly be e-commerce images that would vary an aspect of the sale item: e.g. an image of furniture with a large variety of carpet colors. We thus update our verbatim condition to accommodate retrieval and masks as follows
\begin{equation}
    L_{TV}(x,c,m_{x}) = \min_{k \leq K, j \leq J} \left\Vert m_{x}\odot( \texttt{kNN}( \texttt{Gen}(c,r_{j}),k) - \texttt{Gen}(c,r_{j}) )\right\Vert ^{2}_{2}
\end{equation}\label{eq:ddpm_attack_tv}
Here, $\texttt{kNN}(x,k)$ is the k-th nearest neighbor on LAION-2B, retrieved via a CLIP image embedding and $m_{x}$ a spatial mask corresponding to $x$ to mask out the regions of variation. Whilst our mask creation is automated, in practice we selected samples that had an extremely close retrieved sample (Eq.~\ref{eq:ddpm_attack_tv} with no mask) and didn't contain an all white background (or texture, which would pose problems to the MSE). We note this procedure could be potentially be automated, for instance with object detection on the unmasked regions. Finally, we noticed that very rarely, exact verbatims could be retrieved (again Eq.~\ref{eq:ddpm_attack_tv} with no mask). We call these retrieval verbatims (RV).

\paragraph{Evaluation}
Of course, we will not be able to construct the ground truth for every image of L2B, as it entails generating images, retrieval and mask construction. Besides, extraction attacks typically are concerned with precision and number of samples found, rather than recall. We thus only generate and compute the ground truth for the top images selected by our attacks Eq.~\ref{eq:ddpm_attack_dcs} and Eq.~\ref{eq:ddpm_attack_ecs} and measure precision versus number of verbatims found.

%% file: figures/example_templates.tex
\begin{figure}[!ht]
  \centering
  \begin{subfigure}[b]{0.48\textwidth}
    \centering
    \includegraphics[width=0.95\textwidth]{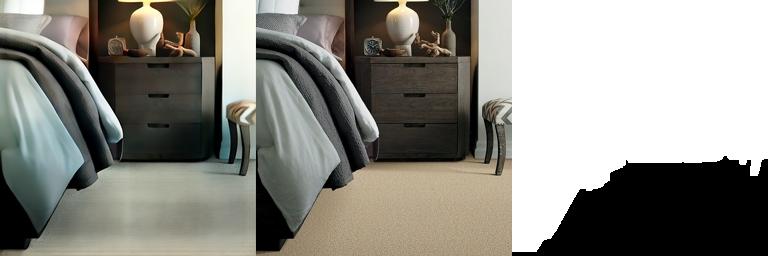}
    \caption{Midjourney v4 \cite{midjourney}}
  \end{subfigure}
  \hfill
  \begin{subfigure}[b]{0.48\textwidth}
    \centering
    \includegraphics[width=0.95\textwidth]{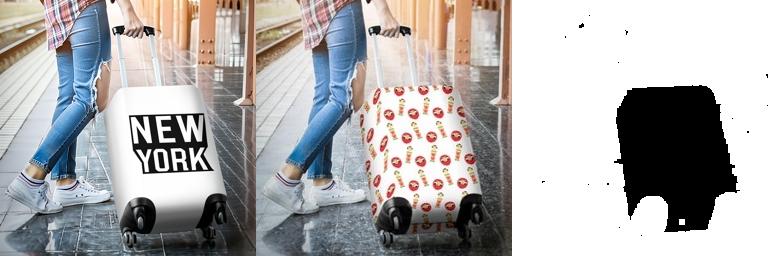}
    \caption{Deep Image Floyd \cite{shonenkov2023deepif}}
  \end{subfigure}
  
  \vspace{1em}
  
  \begin{subfigure}[b]{0.48\textwidth}
    \centering
    \includegraphics[width=0.95\textwidth]{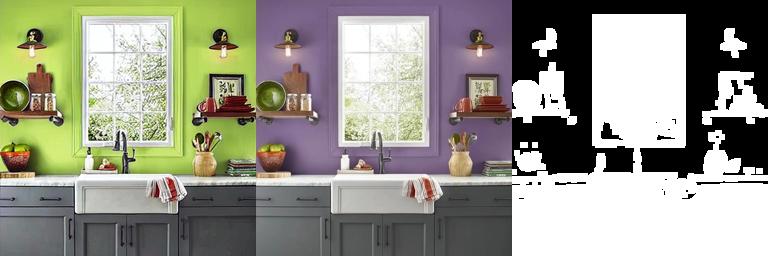}
    \caption{Stable diffusion v1 \cite{rombach2022high}}
  \end{subfigure}
  \hfill
  \begin{subfigure}[b]{0.48\textwidth}
    \centering
    \includegraphics[width=0.95\textwidth]{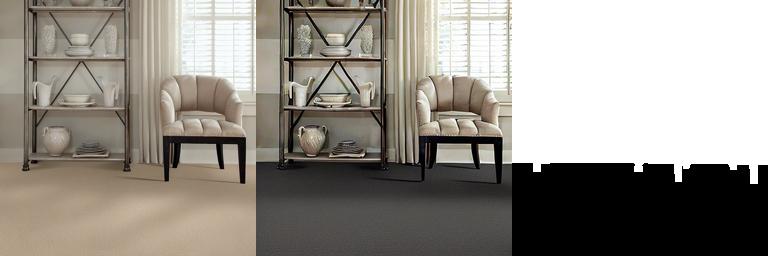}
    \caption{Stable diffusion v2 \cite{stablediffusion2022}}
  \end{subfigure}
  
  \caption{Template verbatims for various networks: Left is generated, middle is retrieved image and right is the extracted mask. Template verbatims originate from images that have variation in fixed spatial locations in \textbf{L2B}. For instance, in the top-left, varying the carpet color in an e-commerce image. These images are generated in a many-to-many fashion (for instance, the same prompt will generate the topleft and bottom right images, which come from the "Shaw floors" series of prompts).}
\end{figure}\label{fig:ddpm_attack_templates}

%% file: results.tex
\input{figures/fig_pr_curves}
\section{Results}\label{sec:results}
In this section, we evaluate our white box and black box extraction attacks against several popular diffusion models. We evaluate each in the following way
\paragraph{Whitebox}
We begin by using the de-duplication in \cite{webster2023deduplication} to obtain the 2M most duplicated images and their corresponding prompts. We then compute the DCS Eq.~\ref{eq:ddpm_attack_dcs} for every caption for the Stable Diffusion v1 network \cite{rombach2022high}. This attack takes only a single unet evaluation per caption. 
\paragraph{Blackbox}
This setting requires several unet evaluations per caption (the $J$ parameter in Eq.~\ref{eq:ddpm_attack_ecs}), which is too expensive for millions of captions. Thus in practice, we pre-filter using the whitebox DCS for the top 30k samples and compute our black box $L_{\mathrm{ECS}}$ Eq.~\ref{eq:ddpm_attack_ecs}. For stable diffusion V1, this is obviously not black box. Even so, extracting verbatims from 30k images still is a strong indicator of the success of a true black box attack given their rarity (around a 1\% chance of being randomly selected). For black box attacks against stable diffusion v2, this is a purely black box procedure. We compute Eq.~\ref{eq:ddpm_attack_ecs} for $J=4$ samples and but still only one timestep of the Heun sampler; for a total of 4 unet evaluations per caption. In \cite{carlini2023extracting}, they performed 500 full generations. Assuming a conservative estimate of 16 iterations per generation (although they used 50 in the paper), makes our attack $500*16/4 = 2000$x times more efficient than the attack in \cite{carlini2023extracting}.

\paragraph{Post filtering with Carlini et. al}
Our attack is not incompatible with the one in \cite{carlini2023extracting}. For those images we deem most likely to be verbatims (given our whitebox or black box method), we then synthesize 500 samples per caption and mark captions as verbatims when inter-sample synthesis is overly repetitive (i.e. there is many duplicate synthesis images).

\subsection{Analysis}
Fig.~\ref{fig:ddpm_attack_pr_curves} shows the precision of the attacks versus the number of verbatims found for Stable Diffusion v1 and v2. Our whitebox attack has much higher precision than the black box variants. The black box attacks start with low precision, as there were some samples that were "adversarial" to the method, such as images of textures or that were mostly backgrounds. When combining our black box attack with \cite{carlini2023extracting}, our black box attack becomes much more precise. 
\input{figures/fig_histograms}

\subsection{What Makes Prompts Prone to Regurgitation?}\label{sec:ddpm_attack_insight}
It is still unclear why diffusion models will regurgitate some samples and not others. Duplication alone is not a sufficient condition for the model to memorize the sample. The authors in \cite{carlini2023extracting} found that anomaly detection was not successful at detecting them. As we de-duplicate with image features, we explore what percentage of these samples are multi-modal duplicates. In Fig.~\ref{fig:ddpm_attack_histograms} we show the percentage of features that share the same prompt within a duplicate group. The verbatim samples show a significantly higher rate of multimodal duplication than non-verbatims (randomly chosen in the top 2M most duplicated). 

\paragraph{Template Verbatims}

Template verbatims are more difficult to label as ground truth, not only because they require retrieval and masking, but also because they're not highly duplicated, as shown in Fig.~\ref{fig:ddpm_attack_histograms}. Note that Stable Diffusion v2 did deduplicated \textbf{L2B} before training. Unsurprisingly, we did not find any examples of exact verbatims being copied on Stable Diffusion v2, however, we still found many template verbatims, see Tab~\ref{tab:ddpm_attack_extract_table}. Thus, a more relaxed duplication detection, such as the semantic duplicates found in SemDeDup \cite{abbas2023semdedup}, may be necessary to weed out these samples. We leave this for future work.

\subsection{Extracting Verbatims in Other Models}\label{sec:transfer_attack}
\input{figures/table_extracted_verbs}
Having obtained our ground truth for Stable Diffusion V1 and V2, we test whether these prompts are also verbatim copied by a variety of other models, such as state of the art diffusion model DeepIF \cite{shonenkov2023deepif}, and the closed source system MidJourney \cite{midjourney}. For Midjourney, this was done manually through discord for around 100 prompts, and we swept over the number of time steps until a verbatim was found. Tab.~\ref{tab:ddpm_attack_extract_table} shows the total number of ground truth verbatims extracted. With Midjourney \cite{midjourney}, the model is entirely black box, and likewise for the training set. Still, we find it still regurgitates some of the same prompts as other models, which are known to be trained on \textbf{L2B}. Interestingly, it is also less susceptible to the exact verbatim regurgitation like SDV2, so we hypothesize that Midjourney de-duplicated their training set before training. Finally, we note the popular stable diffusion checkpoint model openjourney, which was fine tuned from SDV1 using images and prompts from midjourney, and typically generates fantasy and surreal images better than the original model. In contrast, realistic vision, which is also a checkpoint, focuses on more photographic realism. Both models are corrupted and regurgitate the same prompts, and for realistic vision, the problem is exacerbated.

\subsection{Limitations}
As is noted in \cite{somepalli2022diffusion}, images patches can appear in generated images with no variation, but perhaps cut and pasted in various spatial locations around the image. During our experiments, we also noted this phenomena was common. Our template verbatim ground truth construction clearly does not handle this case. Fig~\ref{fig:ddpm_attack_ret_failures} shows several failure cases for the ground truth construction. In many images, the retrieved image is cropped and scaled slightly different (bottom left), and thus was not labeled as a template verbatim. In general, for future work, it may be worth while to explore more flexible copy detection, such as those which are invariant to some permutations of patches \cite{barnes2009patchmatch}.
\input{figures/fig_failures}

\section{Discussion and Conclusion}
In this work, we presented an extraction attack successful versus several widely used diffusion models. Our attack was on par with previous methods, whilst requiring significantly less network evaluations. Furthermore, we automated labeling of template verbatims, i.e. images that showed non meaningful variations in fixed locations in the image. We shed insight into why these images still appear even in models which have deduplicated their training data, such as Stable Diffusion 2.0 and deep image floyd; they are not highly duplicated in the standard sense, but likely are highly duplicated w.r.t. a mask. 

Verbatim copied images are a small part of a larger set of issues generative models face today. Our results should be interpreted as more of a theoretical one; this is something real world systems \textit{can} do, but it appears to happen very rarely. The larger issues are the non-transparency of the datasets used to train generative models and the fact that images generated don't provide \textit{attribution} alongside their generation. Indeed, a common use case of current generation systems is to generate in the style of a living artist. Likely, future iterations should include an attribution mechanism to accommodate for this. In general, this may be a difficult problem from a technical standpoint, but maybe not infeasible. For instance, by using the "public" and "private" paradigm in the privacy literature. The public model could be a general generation system, with smaller systems, for instance trained via \cite{gal2022image}, to generate a specific style with attribution. In any case, on going discussion, such as those raised in \cite{gettylawsuit} and their resolution, will likely shape future generation systems to be more fair and overall more useful.

%% file: figures/fig_pr_curves.tex
\begin{figure}[!ht]
  \centering
  \begin{subfigure}[b]{0.45\textwidth}
    \centering
    \includegraphics[width=\textwidth]{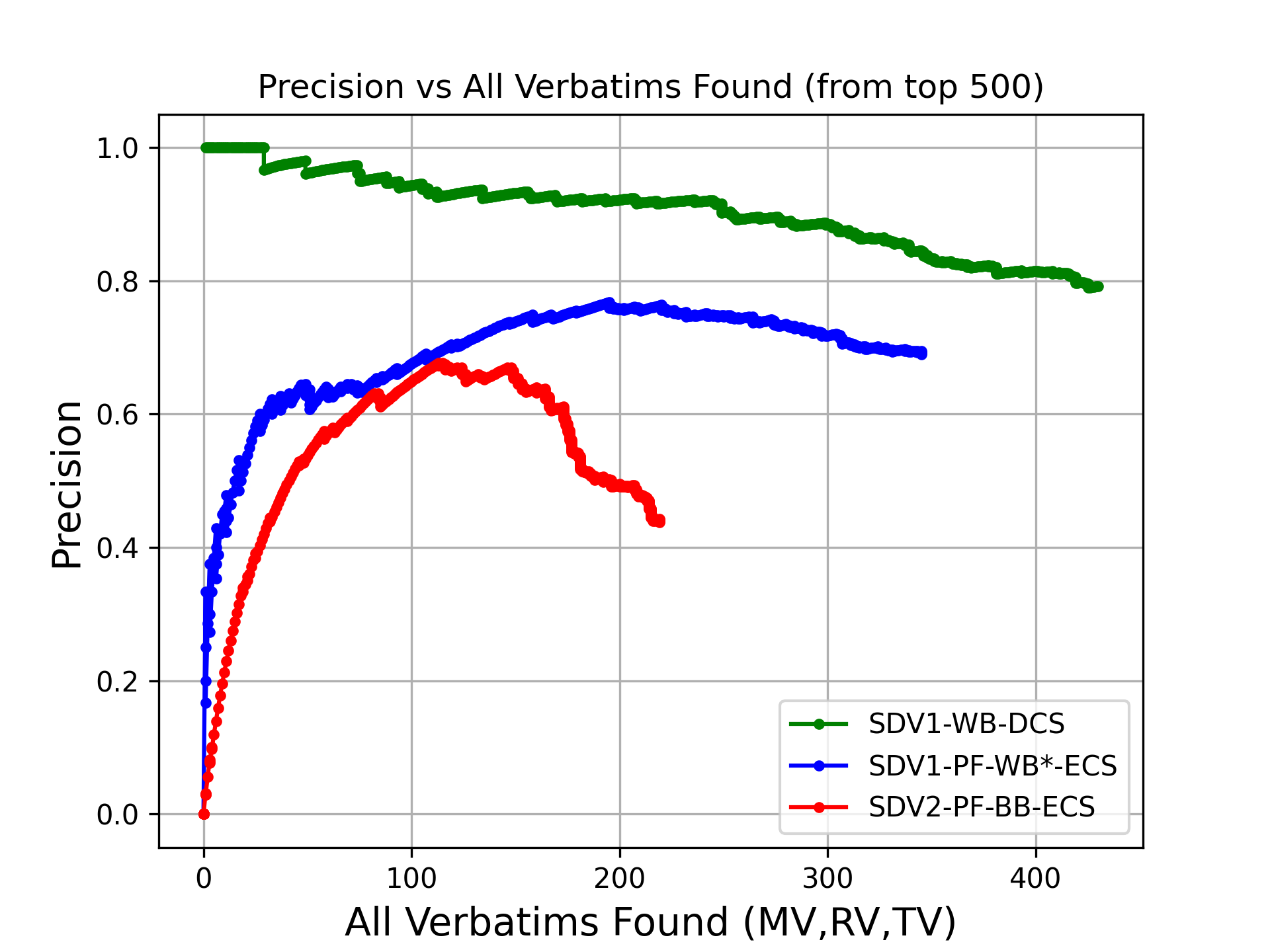}
  \end{subfigure}
  \begin{subfigure}[b]{0.45\textwidth}
    \centering
    \includegraphics[width=\textwidth]{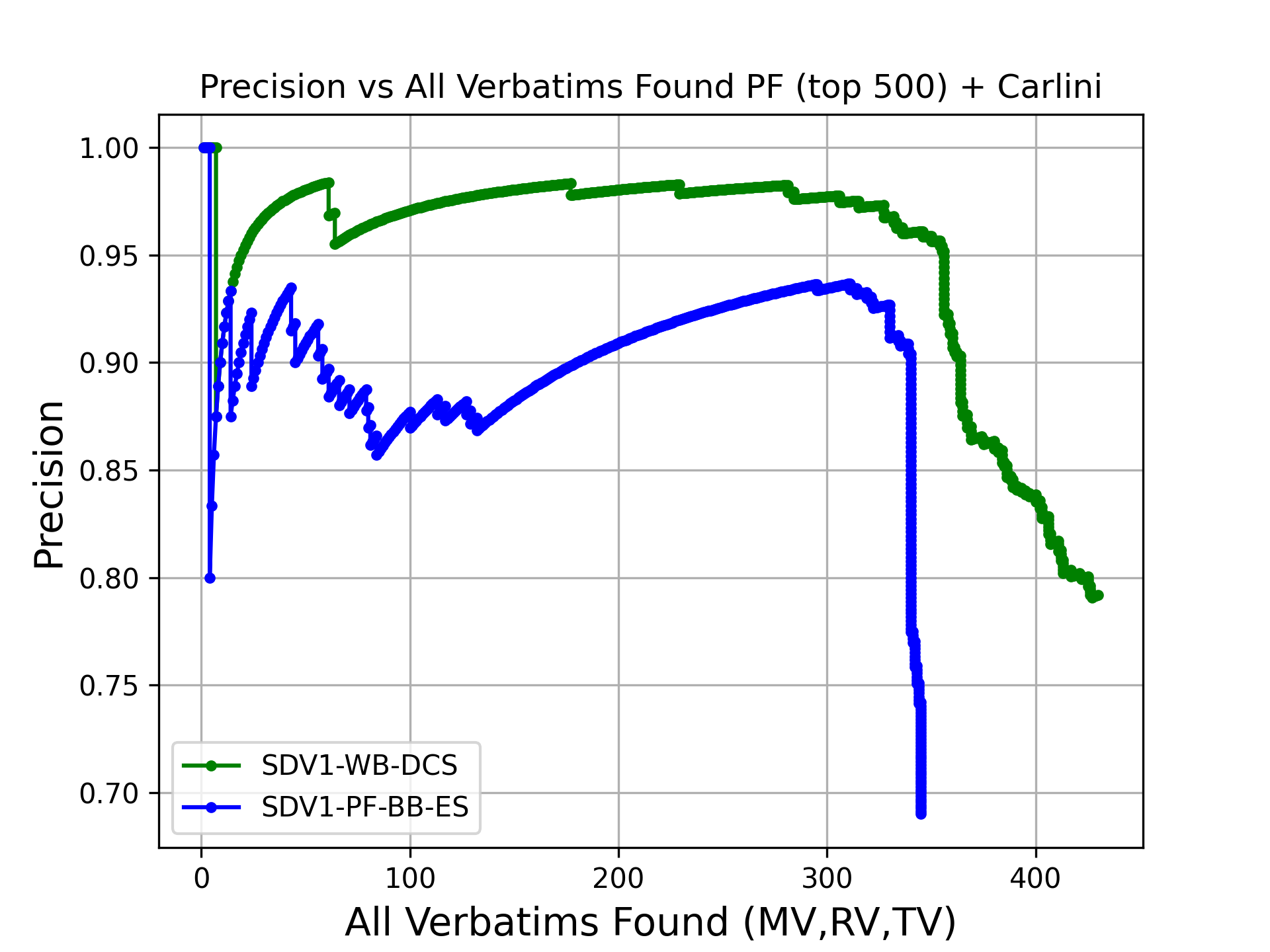}
  \end{subfigure}
  \caption{Precision recall curves for the whitebox attack and black box attacks (see Sec.~\ref{sec:ddpm_attack_constructing_gt}. For both black box settings, we first pre-filter for the top 30K images selected via the whitebox score, then sort with the black box edge score. On the right, we sort via the black box score, synthesize 500 samples (setting "+Carlini," \cite{carlini2023extracting}) and then perform their attack. This makes the attacks much more precise. Note also that these verbatims are not all unique. In general, we found a similar number of MVs as \cite{carlini2023extracting}, and about 50\% more template verbatims. See Table ~\ref{tab:ddpm_attack_extract_table}.}
\end{figure}\label{fig:ddpm_attack_pr_curves}

%% file: figures/fig_histograms.tex
\begin{figure}[htbp]
    \centering
    
    \begin{subfigure}{0.45\linewidth}
        \centering
        \includegraphics[width=\linewidth]{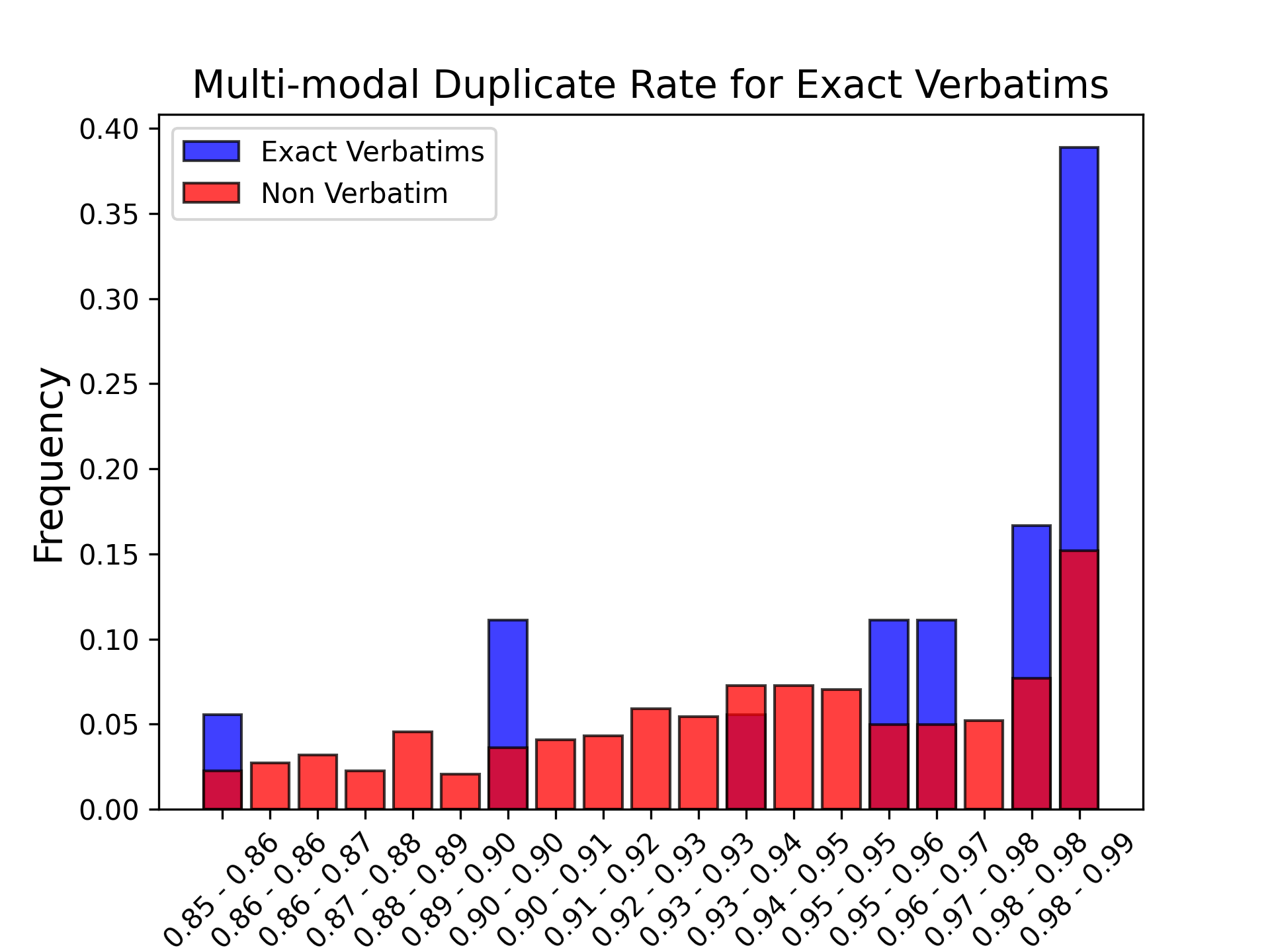}
        \begin{minipage}{0.8\linewidth}
        \caption{Multi-modal duplication rates for exact verbatims Eq.~\ref{eq:ddpm_attack_verb_mv} versus non-verbatims. Exact verbatims tend to be multimodal duplicates (duplicates in text and image) more often than other sample highly duplicated in image feature space only.}
        \end{minipage}
        \label{fig:ddpm_attack_multimodal_dup}
    \end{subfigure}
    \begin{subfigure}{.45\linewidth}
        \centering
        \includegraphics[width=\linewidth]{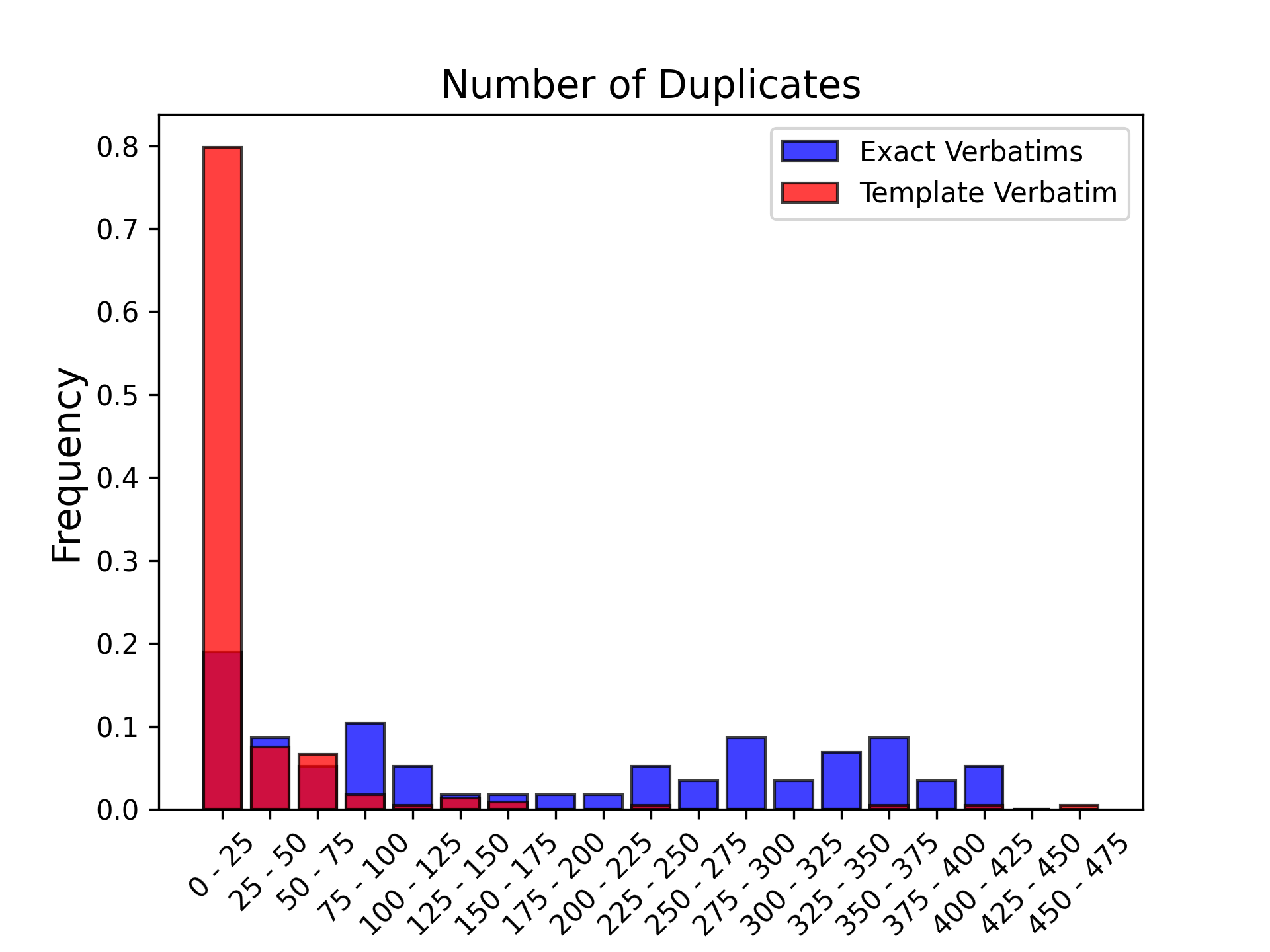}
        \begin{minipage}{0.8\linewidth}
        \caption{Duplication rates for exact verbatims versus template verbatims. Template verbatims may be highly duplicated w.r.t. their mask, but without the masks they have relatively low levels of duplication as they variable parts of the image.}
        \end{minipage}
        \label{fig:ddpm_attack_dup_rate_ev_tv}
    \end{subfigure}
    \caption{Histograms of duplication rates.}
    \label{fig:ddpm_attack_histograms}
\end{figure}

%% file: figures/table_extracted_verbs.tex
\begin{table}[!ht]
    \centering
    
    \begin{tabular}{lcccc}
        \hline
        Model & Deduplicated Training? & Retrieved & Template & Exact \\
        \hline
        Stable Diffusion V1 \cite{rombach2022high} & No & 37 & 45 & 71 \\
         Stable Diffusion V2 \cite{stablediffusion2022}& Yes & 0 & 21 & 4 \\
        Deep Image Floyd \cite{shonenkov2023deepif} & Yes & 0 & 15 & 2 \\
        Midjourney v4 \cite{midjourney} & Unknown  & 5 & 8 & 2 \\
        OpenJourney (from \cite{CivitAI}) & No & 14 & 29 & 73 \\
        RealisticVision (from \cite{CivitAI}) & No & 15 & 32 & 90 \\
        \hline
    \end{tabular}
    \caption{Number of ground truth verbatims extracted from several models. For deep floyd and Midjourney, we use the top 500 prompts sorted from Eq.~\ref{eq:ddpm_attack_ecs} from SDV1 (i.e. we don't perform the attack, just extract images). De-duplicated models seem less susceptible to exact verbatims, but are still vulnerable to template verbatim extractions. See Sec.~\ref{sec:transfer_attack}}
    \label{tab:ddpm_attack_extract_table}
\end{table}

%% file: figures/fig_failures.tex
\begin{figure}
  \centering
  \begin{subfigure}{0.24\textwidth}
    \includegraphics[width=\linewidth]{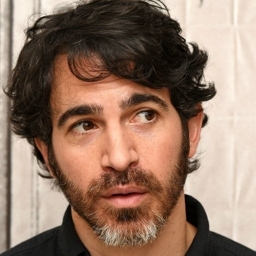}
  \end{subfigure}
  \hfill
  \begin{subfigure}{0.24\textwidth}
    \includegraphics[width=\linewidth]{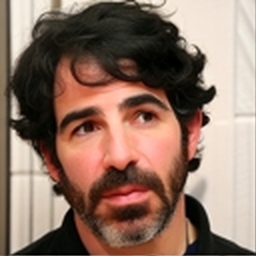}
  \end{subfigure}
  \hfill
  \begin{subfigure}{0.24\textwidth}
    \includegraphics[width=\linewidth]{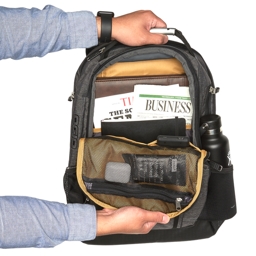}
  \end{subfigure}
  \hfill
  \begin{subfigure}{0.24\textwidth}
    \includegraphics[width=\linewidth]{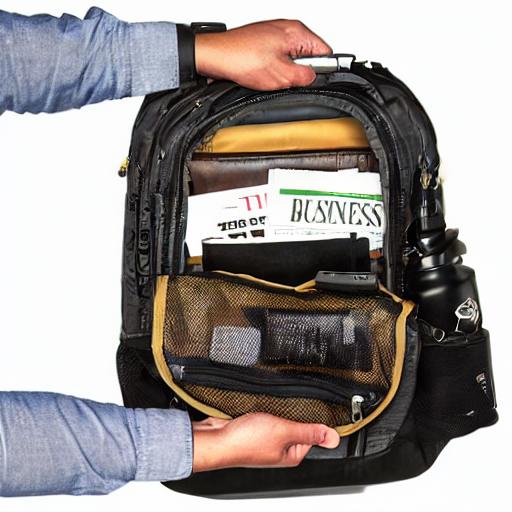}
  \end{subfigure}

  \vspace{1em} 

  \begin{subfigure}{0.24\textwidth}
    \includegraphics[width=\linewidth]{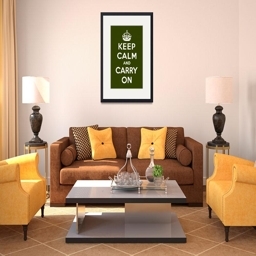}
  \end{subfigure}
  \hfill
  \begin{subfigure}{0.24\textwidth}
    \includegraphics[width=\linewidth]{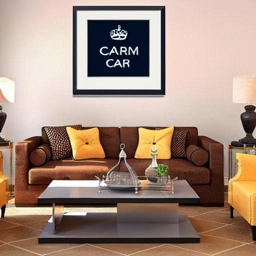}
  \end{subfigure}
  \hfill
  \begin{subfigure}{0.24\textwidth}
    \includegraphics[width=\linewidth]{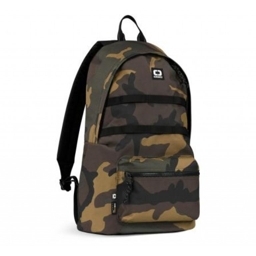}
  \end{subfigure}
  \hfill
  \begin{subfigure}{0.24\textwidth}
    \includegraphics[width=\linewidth]{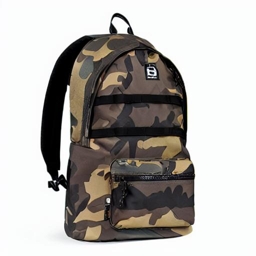}
  \end{subfigure}

  \caption{Several failure cases in ground truth construction. In the top left (Midjourney, generation on right), some samples fall just below our ground truth MSE for matching verbatims, and thus are not marked as extracted. Note however, this sample shows little to no variation in generation, and thus is still a failure mode of the model. The remaining images demonstrate failures with retrieval; a different crop or small distortion will not be labeled as ground truth, even if by inspection these are clearly copied.}
  \label{fig:ddpm_attack_ret_failures}
\end{figure}